# Can AI be Accountable?

Andrew L. Kun
University of New Hampshire
andrew.kun@unh.edu

# 1

# Can AI be Accountable?

## Summary

The AI we use is powerful, and its power is increasing rapidly. If this powerful AI is to serve the needs of consumers, voters, and decision makers, then it is imperative that the AI is accountable. In general, an agent is accountable to a forum if the forum can request information from the agent about its actions, if the forum and the agent can discuss this information, and if the forum can sanction the agent. Unfortunately, in too many cases today's AI is not accountable – we cannot question it, enter into a discussion with it, let alone sanction it. In this chapter we relate the general definition of accountability to AI, we illustrate what it means for AI to be accountable and unaccountable, and we explore approaches that can improve our chances of living in a world where all AI is accountable to those who are affected by it.

## 1.1 What is Accountability? Why does it matter in AI?

Accountability is a key element of good governance. Thus, we can find examples of governmental accountability throughout history – from the Chinese empires (Hucker, 1951; Guy, 2021), to pre-colonial West African tribes (Palagashvili, 2018), to the medieval kingdom of England (The National Archives), to the European Union (Bovens, 2007). Here we are concerned with accountability for AI, but before we turn our attention to AI, let us look at one of these examples, the 13[th] century effort to introduce accountability for the king of England through the Magna Carta (The National Archives).

In 1215 the Magna Carta constrained the powers of the king of England by making him accountable to an elected body. Importantly, the Magna Carta, some eight centuries ago, included an explicit mechanism for accountability of the king. Accountability has three constituents parts – the ability of a forum to request information from an agent about the actions of that agent, the ability to discuss the information provided by the agent, and the ability to sanction the agent if the forum finds this necessary (Mulgan, 2000; Bovens, 2007; Brandsma and Schillemans, 2013; Lindberg, 2013; Wieringa, 2020; Novelli et al., 2024). The following two sentences of Clause 61 of the Magna Carta address these three constituent parts:

> "The barons shall elect twenty-five of their number to keep, and cause to be observed with all their might, the peace and liberties granted and confirmed to them by this charter."

> "If [the king or his officials] offend in any respect against any man, or transgress any of the articles of the peace or of this security, and the offence is made known to four of the said twenty-five barons, they shall come to [the king or the chief justice] to declare it and claim immediate redress."

The first sentence identifies the forum which can hold the king, that is the agent, accountable. The forum is an elected body of 25 barons. The first sentence is also quite direct regarding sanctions: if the forum determines that the king is encroaching on their rights, they can "cause [peace and liberties] to be observed with all their might," that is with military force. In other words: encroach on our liberties and we will hold you accountable by sending military force against you.

The second sentence describes the process for questioning the king and discussing the king's actions. This questioning can be initiated by any four of the 25 elected barons. These four barons can question the king or chief justice, and "claim immediate redress." It is understood that, if the king does not provide an explanation, and does not correct the problem, then the sanction, or consequence, is the use of military force, mentioned in the first sentence.

### 1.1.1  AI and accountability

To identify the concept of accountability, let us start with a problem statement. In general, when an agent undertakes actions, these actions affect others. Thus, the actions of the agent are of interest to others – from the motivation, to the mechanism of completing the actions, to consequences of those actions. For example an AI can be the agent that executes actions, such as making financial decisions (e.g. who can get a loan from a bank), controlling physical devices (such as the trajectory of an automated vehicle), controlling organizational decisions (such as an AI scheduling first responders for shifts), or providing advice (such as an AI responding to a query about which coffee machine the user should buy). The *problem* is that AI is powerful and inscrutable, and the people who are affected by the actions of AI lack protection from those actions.

We can compare this to the problem addressed by the Magna Carta. Before 1215 the king of England tried to take absolute power into his hands, and his actions were very much of interest to the landowners around him. But the problem for the landowners was that the mechanisms for questioning, discussing, and sanctioning the king's actions were either nonexistent or at least weak. Today, the problem we face as consumers, voters, and decision makers, is that the mechanisms for understanding, discussing, and sanctioning the actions of different AIs are either nonexistent or weak. As AI gets more powerful, both computationally, and with its ability to affect physical systems, this problem is likely to get worse and more important. And this is what connects us as users of AI to the barons of the Kingdom of England in 1215: if actions by an agent are inscrutable and unchangeable, then individuals affected by those actions cannot be protected from the agent. The English barons needed protection from their king. And 21st century humans, from consumers, to voters, to decision makers, need protection from AI.

With this problem in mind, we can identify a *goal*: we need a system that will protect users from the actions of an AI. The system should compel the parties responsible for developing and operating the AI to keep the interests of the users front and center.

*How* can this system be implemented? One answer is: accountability. Bringing the above ideas together, we can describe  accountable AI as follows. Users of AI need protection from a possibly powerful, yet often inscrutable tool - AI. The goal is to protect users, at least with improvements over longer time periods (e.g. as new releases of a software appear). The way to accomplish this is by setting up an enforceable mechanism of answerability – that is an accountability mechanism – where an appropriate forum can pose questions about the AI to a responsible party, render judgment about the actions of the AI and effect positive or negative sanctions, which can lead to updates in the AI (again over the long run). The forum, as well as the responsible party (or agent), can take different forms. For example, a manager can act as the forum for AI developers through a review; citizens can hold policymakers accountable for their AI-related policies through elections; stakeholder groups can initiate audits of AI developers, and publicize the results to hold the developers accountable; and government regulators can hold an AI company accountable through audits and regulations (Novelli et al., 2024; Birhane et al., 2024; Wilson et al., 2021).

### 1.1.2 The Accountable AI Markov chain

From the preceding text we can see that the basic idea of accountability for AI is fairly straightforward: users need protection from a possibly powerful AI. But, the implementation of accountable AI is a complex issue. Different researchers, practitioners, and policymakers use different frameworks and definitions to bring it to life (Novelli et al., 2024). In order to describe and compare different approaches to, and representations of, accountability for AI, we introduce the Accountable AI Markov chain, shown in Figure 1.1. The Markov chain consists of states (shown as circles), connected by possible transitions between states (the arrows) (Rabiner, 1989). Since there are multiple possible transitions from some states, this means that each transition can occur with some probability.

Let us now review the Accountable AI Markov chain. Starting on the left, the first state of the Markov chain is "Delegate task to AI agent." This state represents the idea that a human user delegates to the AI the execution of some task. This state is also a simplified representation of all of the steps necessary to design, deploy, and maintain the AI agent. The next state in the chain denotes the usual start of the accountability process. It is the "request to explain, justify to forum," which is followed by the "response to forum," and then "debate, discussion." These three states are at the core of the interaction between the forum and the agent. Although, as we shall see, when it comes to accountable AI, the focus is usually on the request for information and the response of the agent (e.g. (Sandvig et al., 2014)). Debate and discussion between the forum and the agent, or a party responsible for the agent, is often not mentioned (cf. (Brandsma and Schillemans, 2013)).

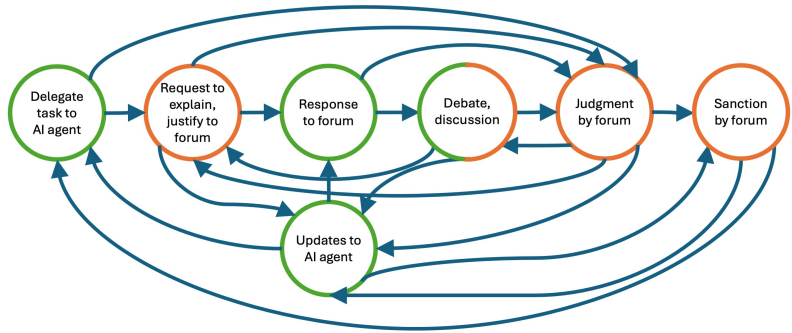

Figure 1.1 The Accountable AI Markov chain represents the AI account- ability process as Markov states connected by transition probabilities. States associated with the AI are represented by green circles. Stated associated with the forum are represented by orange circles. The "Debate, discussion" state is associated with both AI and the forum, so its circle is two-colored.

Moving through the chain, we reach the "judgment by forum," which is an evaluation of the actor. If the evaluation is incomplete, it can be followed by additional debate and discussion, or another round of requests for data. Once the judgment is final, the next state is "sanction by forum." "Sanction" is a critical outcome of the accountability relationship between the forum and the agent. Understanding the motives and actions of an AI is not enough to counteract the power imbalance between AI and users (Longo et al., 2024). If there is no sanction, then there is no accountability. The sanction, either for refusing to respond to a request for information[1], or as a result of the judgment, is what can compel the actor to engage in the process, accept judgments, and make changes that support users (Wieringa, 2020).

The last state in Figure 1.1 is "updates to AI agent." This state is usually not mentioned when discussing accountability, and we could consider it to (at least sometimes) be part of the "sanctions" state. However, in this model we separate "updates" to clarify that accountability should be about the future, and not just the past (Bovens, 2007). Yes, agents should be accountable for their past actions. But, ultimately, the problem that accountability is supposed to solve is that users need protection from harmful and inscrutable actions by AI. The goal of accountability is not to reward or punish past behavior; those are simply the tools that accountability uses to accomplish its actual goal, which is to steer actors towards good future behavior. This is why forum judgments and/or sanctions should lead to updates in AI.

Bovens (2007) captures the "thin line" between the retrospective asking of questions and the prospective changing of policies in government, writing that "accountability is not only about ex post scrutiny, it is also about prevention and anticipation... Many actors will anticipate the negative evaluations of forums and adjust their policies accordingly. Thus, ex post facto accountability can be an important input for ex ante policy making."

Ideally, a similar "thin line" will separate the scrutiny of an algorithm by a forum, from updates to that algorithm that will bring it closer to performing according to the forum's standards. Neyland (2016) describes exactly this dynamic in a paper that reports on the development of an accountable video surveillance system (also quoted by (Wieringa, 2020)):

---

[1] Lindberg (2013) argues that the forum sanctioning the agent for not responding (or responding inadequately) to information requests is the only type of sanction that is part of accountability.

"Accountability was not accomplished in a single moment, by a single person, but instead was distributed among project members and the ethics board and across ongoing activities, with questions taken back to the project team between meetings and even to be carried forward into future projects after the final ethics board meeting."

The model in Figure 1.1 represents these possibilities by including transitions from the "request" state, as well as from the "debate" state, to the "updates" state.

## 1.2  Why is Accountability challenging in AI? Does it create conflicts with other Responsible AI principles? What makes it so hard to solve?

Accountability manages information asymmetry between the forum and the agent (Brandsma and Schillemans, 2013). The agent has information about the delegated task, while the forum can often only gain detailed information through the cooperation of the agent. Accountability is in place to limit the power of the agent, and make it more likely that the agent will follow the desires of the forum. This goal of limiting power, and the information asymmetry, lead to two challenges for accountability of AI.

One challenge is the same for AI users as it was for the barons in 1215 England: uneven distribution of power. In 1215 the King of England was quite powerful, and ready to assert his power over others. Today, AI provides many individuals and organizations with very significant powers, and given their power they might simply refuse to be held accountable. Now, in many industries accountability is part of doing business. For example, companies producing computer networking equipment cannot refuse to have their products tested for compliance with networking standards, because of the power of market pressures[2] – consumers will not purchase a WiFi hotspot that works with some WiFi–enabled devices but not others. Similarly, car companies cannot refuse to have their vehicles inspected by state-approved mechanics whose job it is to confirm that the vehicle is road-worthy, because of the power of legal requirements (Inners and Kun, 2017). But neither market pressures nor legal requirements are powerful enough to reliably make today's AI products accountable.

---

[2]  Market pressures bring networking companies to the industry-funded Interoperability Lab at the University of New Hampshire, where they can confirm the interoperability of their products – see https://www.iol.unh.edu.

The second challenge for accountable AI is how to overcome the information asymmetry. Here we face the fact that much of our AI works as a black box, and the inner workings are not understandable even to professionals, let alone to users of AI (e.g. the military personnel using autonomous weapons systems (Chavannes et al., 2020)), or to lawyers, politicians, and the general public.

Even worse, the problem we face is that AI can act like the evil demon that worried the French philosopher Descartes. Descartes argued that our senses often deceive us, and worried that an evil demon might take possession of his senses and make the world appear completely different to him than it really is (Descartes, 2005). The ultimate version of such an evil-demon-like AI would be a simulation – if we, sentient agents, live in a simulation (see e.g. Chalmers (2022)), then the AI that is behind the simulation is clearly unaccountable to us: we cannot question it or judge it. But there is no need to go as far as to assume that we are in a simulation. Unfortunately, today's AI can indeed take possession of our senses, just as Descartes feared the evil demon would. From deep fake videos that purport to show events that never happened, to bots that pretend to be human users on social media, there are many AI "demons" that actively distort our view of reality. They refuse accountability: when interrogated, they actively misrepresent who they are, what they are doing, and why. That is, when a forum queries them, they lie about who the agent is, about the true nature of their actions, and about their motivations. They fundamentally deceive us about who is optimizing which outcomes. For example, we might believe that the motivation of a social media site is to allow users to reach unfiltered information, and that the site is therefore maximizing information flow from such unfiltered sources. But, the truth might be that the motivation of the site is to promote a product or idea, and the site is optimizing the flow of information of one particular tilt. In the words of Mitchell et al. (2025), we cannot trust AI to be truthful without accountability.

Furthermore, today's AI can be powerful in the hands of people who otherwise do not wield conventional power. Computing and algorithmic power has been democratized, which means that, to paraphrase Suleyman (2023), one doesn't have to be king of England to wield AI power. With so many potential developers and operators of AI, implementing AI accountability is becoming more and more difficult.

Finally, the power of AI is increasing. Today's AI is not simply providing answers to our queries. Rather, as Harari (2024) argues, it is an alien intelligence which can often act as an agent in its own right – not an agent with consciousness or feelings, but one with considerable independence and power nevertheless. And it might become powerful enough to evade our attempts at making it accountable.

While accountability for AI is challenging, it does not in general conflict with other responsible AI principles. It can support fairness, safety, sustainability, and lawfulness. It can be supported by explainability, transparency, auditability, and contextualization. Still, conflicts can exist. For example, information disclosure for accountability could expose information about the system that can create a conflict with safety requirements, with legal protections for intellectual property, as well as with privacy requirements. Furthermore, conflicts between responsible AI principles can occur as a result of different stakeholders prioritizing different AI values. For example, Jakesch et al. (2022) demonstrate how AI practitioners can have different priorities than the population at large. They conclude that it is important to "[pay] attention to who gets to define what constitutes 'ethical' or 'responsible' AI."

## 1.3  How is Accountability applied in practice? What methods or tools are used?

### 1.3.1  Auditing AI

For accountability to work, a forum must be able to gain insight into the actions of the actor. Sandvig et al. (2014) proposed that this can be done through algorithm audits. This proposal builds on so-called audit studies that were designed to identify discrimination. In an audit study two matched testers who are highly similar but differ clearly in one aspect (e.g. ethnicity) interact with an actor (e.g. a landlord) to identify any differences in treatment; many such interactions between matched testers and actors are recorded to identify possible discriminatory trends (Riach and Rich, 2002). Inspired by audit studies Sandvig et al. proposed multiple approaches to auditing algorithms, from direct review of the code, to designs where real or simulated users interact with an algorithm, and thus gather data on how the algorithm behaves under different circumstances.

In the last decade AI has grown by leaps and bounds, and so have AI auditing tools. Ojewale et al. (2024) conducted interviews with experts and reviewed 435 audit-related tools. The authors categorize tools as supporting one or more of seven proposed stages of the audit process. These proposed stages are:

1. **Harms discovery:** An audit must first identify that there is an AI system to examine, since this is not always clear. The identification of the AI system should also include the identification of potential harms by the system.
2. **Identifying principles and norms to guide the audit:** In a number of domains, from government, to banking, to healthcare, audits follow very clear guidelines (Raji et al., 2020; Birhane et al., 2024). A category of AI auditing tools attempts to provide the same for AI audits.

3. **Gathering information through transparency infrastructure:** Some AI tools are accessible to auditors because they implement a transparency infrastructure, e.g. by publishing an API. When this is the case, audit tools can help auditors query AI as part of an audit.

4. **Gathering information through external data collection:** When the algorithm cannot be queried directly, AI can only be assessed through information that is gathered about its behavior.

5. **Performance analysis:** Once data is gathered, tools can help assess how an AI stacks up regarding accountability – these tools can e.g. calculate scores that are identified in the "principles and norms" step above. The use of one such score by Wilson et al. (2021), the four-fifths rule, is discussed in section 1.4.1 below.

6. **Audit communication:** The audit supports accountability, and accountability must include sanctions. Ojewale et al. (2024) argue that a prerequisite for sanctions is having audit results that can be understood by a broad audience. Audit communication tools help auditors present data in a widely-understandable way.

7. **Audit advocacy:** The authors present "audit advocacy" as the sanction stage: the auditors will inform an audience of the results, and this can enable change in how AI is created, deployed, and operated.

One interesting question is this: how do the seven auditing stages proposed by Ojewale et al. (2024) map onto the Accountable AI Markov chain in Figure 1.1? The answer to this question is shown in Figure 1.2. We can argue that "harms discovery" maps loosely onto the "Delegate task to AI agent" state. Note that for Ojewale et al. the goal of harms discovery is to "identify and prioritize audit targets and harms to investigate," and this requires identifying a deployed AI. The "principles/norms" stage does not neatly map onto one or two states – rather it informs all of them. The "gathering information" stages (either using the transparency infrastructure or external data) map onto the forum's "request to exlain, justify" and any subsequent "response" state in the Markov chain. "Performance analysis" maps onto the "debate and discussion," as well as the "judgment by the forum" states. Finally "communication/advocacy" maps onto the "sanctions" state. There is no stage that neatly maps onto the "updates to the AI" state, although that would be a result of any sanction activity. Thus, the seven stage model of Ojewale et al. (2024) broadly matches our Accountable AI Markov chain model, although the seven stage model is explicit in addressing principles and norms, while it does not explicitly address debate and discussion or updates to the AI.

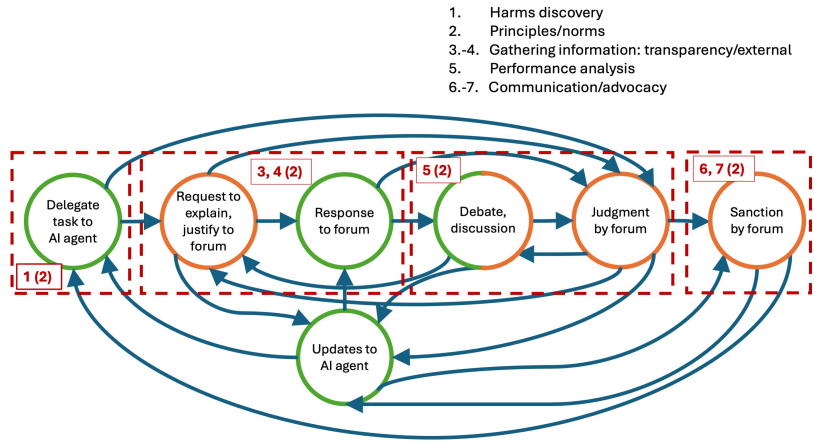

Figure 1.2 The seven stages of the audit process by Ojewale et al. (2024) mapped onto the Accountable AI Markov chain. The two models broadly agree, however, Ojewale et al. explicitly addresses principles and norms, while they do not explicitly address debate and discussion or updates to the AI.

Birhane et al. (2024) review audit methods and practices aimed at accountability for AI. The authors provide a rich source of real-world AI audit approaches outside of academia. These AI audits often result in significant sanctions and changes – one example is the UK Information Commissioner's Office fining TikTok for improper use of children's data (Information Commissioner's Office, 2023). Furthermore, the authors contribute important observations about how these methods and practices can have an impact on AI accountability. One such observation is that audits should focus on more than just evaluation. Referring to Figure 1.2, we can understand the authors' argument to be that accountability requires more than a focus on gathering information and performance analysis. In other words, accountability requires more than an answer to "how do we assess this AI?" Rather, it also requires answers to "why are we assessing this AI?" and "what is the goal of assessing this AI?" In other words, we need a problem statement and a goal statement, along with a description of methods.

Still, while audits can be an important part of the accountability process, Birhane et al. (2024) warn that audits can also be ineffective and harmful; for example audits can reinforce power asymmetries, and, in a practice called audit-washing, they can be used by powerful interests to create the impression of accountability, without supporting true accountability.

### 1.3.2  Guiding the work to create accountable AI

Making AI accountable is a complex undertaking and it can be guided by predefined processes and requirements, including frameworks, guidelines, checklists, standards, and legal regulations.

One *framework* relevant to accountable AI is the Internal Audit Framework, introduced by Raji et al. (2020). We will highlight three elements of this framework:

1. The framework advocates for the use of "datasheets for datasets," which is a collection of information about "motivation, composition, collection process, pre-processing/cleaning/labeling, uses, distribution, and maintenance" of the dataset (Gebru et al., 2021). The aim of the datasheet is to "mitigat[e] unwanted societal biases or potential risks or harms" when using a dataset to train a machine learning model.

2. The framework also includes the use of "model cards for models." A model card provides information about a model, including evaluations of how a *trained* machine learning model performs in different relevant contexts (Mitchell et al., 2019). This information can help in the responsible deployment of an AI.

3. The framework places a focus on selecting those stakeholders who can conduct the audit successfully. This will require technical expertise, skills in assessing potential harms and societal impacts, and also the ability to handle the potential "ambiguous conclusions."

Constantinides et al. (2024) present a list of 22 *guidelines* for developing responsible AI. Their guidelines are grouped into six categories: intended uses, harms, system, data, oversight, and team. Referring to Figure 1.1, the guidelines support the Markov states marked in green – those that have to do with the socio-technical work of making the AI a reality and responding to the forum. For example, Guideline 9 instructs developers to "[p]rovide mechanisms for interpretable outputs and auditing." This can directly support functions for the "Request to explain, justify to forum" state in the Accountable AI Markov chain.

Building on the guidelines proposed by Constantinides et al. (2024), Bogucka et al. (2024) introduce a framework for pre-filling impact assessment reports. Impact assessments can be legally required for AI systems, and as such they will often be part of an accountability process. Yet, they can be challenging and expensive to produce, especially for smaller companies. The framework of Bogucka et al. (2024) supports this work by asking stakeholders questions about the AI system, and using their responses to generate the impact assessment report. The report starts with system information collected from stakeholders. It also includes stakeholder-reported risks, benefits, and risk mitigation approaches. The system expands these three areas with additional possible risk and benefits, as well as with suggested mitigation approaches.

*Standards* that guide work with AI can be broad, providing guidance for a wide range of activities, or relatively narrow, focusing on a specific area of activity. An example of a broad standard is the IEEE Standard for Software Reviews and Audits (IEEE Computer Society, 2008). This standard applies to all software, which includes AI. It provides detailed guidance on how to create documentation for accountability of software. One example of a narrow standard is the Minimum Information about Clinical Artificial Intelligence Modeling for Generative AI (MI-CLAIM-GEN) (Miao et al., 2025). As its name indicates, this is a standard that guides one type of activity: clinical studies that use generative AI. The standard supports accountability because it guides researchers in collecting the minimum amount of data that is deemed necessary for reporting on a clinical study using generative AI. When the researchers are asked by a forum to provide information about their study, they will have that information ready.

With AI developing very fast, policymakers are trying to respond by crafting *regulations*, as well as guiding agreements for AI governance. Examples of this developing landscape include the "Ethics guidelines for trustworthy AI" (High-Level Expert Group on Artificial Intelligence, 2019), the EU AI Act (European Parliament and Council, 2024), the US Algorithmic Accountability Act (United States Congress, 2023), and the intergovernmental OECD recommendations on AI (OECD, 2019).

As for *regulation*, we can expect that the practice of AI accountability will be regulated at different organizational levels. This is already the case with automated driving, where regulation might be international, national, and regional (Inners and Kun, 2017), and maybe even local. An example of a simple local regulation of AI-supported technology is the prohibition of the use of drones near the Great Geysir on Iceland (Figure 1.3). On the trans-national level, the EU AI Act outlines obligations for AI to be used in the EU, if it is classified as limited-risk or high-risk (Thelisson and Verma, 2024). Importantly, high-risk AI must fulfill the conformity assessment legal requirement – this must be done before AI deployment, as well as on a continuing basis afterwards.

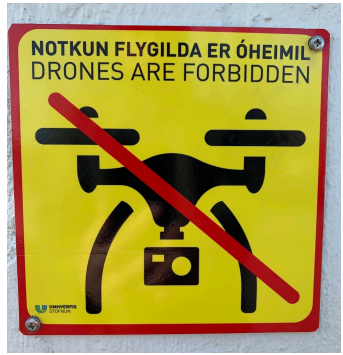

Figure 1.3  A marker near the Great Geysir on Iceland tells visitors that drones are forbidden. This is an example of local regulations already impacting AI use. We can expect that this will continue. Photo by author, 2019.

## 1.4  What's an example of Accountable in AI done well? What's an example of Accountable in AI gone wrong?

### 1.4.1  Accountable AI done well

While the literature presents many examples of problems with AI accountability, one example of accountable AI done well is the case of *pymetrics* presented by Wilson et al. (2021).

At the time Wilson and colleagues conducted their study in 2020, *pymetrics* was a company that supported hiring by client firms. Specifically, *pymetrics* performed candidate screening, and they worked to make their decisions fair. Candidate screening was done using an algorithm that *pymetrics* trained using information from each of their clients – the information pertained to the firm and the job for which they were hiring.

*pymetrics* defined fairness as avoiding two types of discrimination defined in the Civil Rights Act (US Congress, 1964): "disparate treatment," and "disparate effects." Disparate treatment happens when members of different classes of persons are treated differently in a process, such as a hiring process; this is a clear case of discrimination. Operationalizing fairness in algorithms to avoid disparate treatment means not using certain attributes (e.g. racial background) as input features when training models. *pymetrics* operationalized avoiding disparate effects discrimination by using the four-fifths rule of the Uniform Guidelines On Employee Selection Procedures (Equal Employment Opportunity Commission, 1978). This rule defines harm as occurring if one group of applicants is hired at a lower than 80% rate than the group that is hired at the highest rate. *pymetrics* tested each of its models for compliance with this four-fifths rule before delivering it to a client, and they also used longitudinal approaches to improve their modeling.

The audit team set out to assess how well *pymetrics* was able to keep its commitment to fairness. To do this, in the audit they asked the following questions:

1. Is testing for compliance with the four-fifths rule correctly implemented?
2. Is demographic data (mis)used as training input to models?
3. Could a malicious attacker corrupt training such that using the model would result in disparate treatment of different groups?
4. Are there checks in place to prevent human error, or human manipulation, to result in an unfair model being created?
5. Are there assumptions about the data, data processing, or model training, that could trip up the testing for compliance with the four-fifths rule?

For all of these questions the audit team found answers that indicated an adherence to accountability by *pymetrics*. In the words of Wilson et al. (2021) "[w]ith respect to the results of our audit, we are comfortable stating that *pymetrics* passed the audit, subject to the qualifications and limitations we state..."

Importantly, Bovens (2007) points out that the forum should be able to discuss the account of the actor with the actor. To a degree in keeping with this approach, the audit was conducted in cooperation with *pymetrics*. The audit team uses the term "cooperative audit." The cooperation took the form of "onboarding" the audit team by *pymetrics* employees as well as technical assistance with code. This cooperation is similar to the code audit proposed by Sandvig et al. (2014), although Sandvig et al. did not consider such an audit likely, and did not envision additional involvement from the creators of the algorithm in the audit. Importantly, the cooperation also included an agreement that the audit team would give *pymetrics* 30 days to address any negative findings before the findings were made public.

Figure 1.4 shows how the Accountable AI Markov chain that we introduced earlier can describe the audit of *pymetrics*. First, the task was delegated to the  AI algorithm – here this means that *pymetrics* created a model. Next, the audit team requested information, which the *pymetrics* team provided. This was followed by an internal evaluation ("judgment") by the audit team, and without direct inclusion of *pymetrics*. Thus, the "debate" state was skipped. New requests were generated, and this loop was completed more than once. Next, the team provided its judgment. *pymetrics* was given the opportunity to update its AI before the sanction – in this case the sanction was simply the announcement of results. Finally, *pymetrics* AI continued its operation. Note that the collaboration between auditors and *pymetrics* staff echoes the recommendation of the Internal Audit Framework (Raji et al., 2020) that there should be "a mitigation plan or action plan, jointly developed by the audit and engineering teams," to address product shortcomings that are revealed in an internal audit. We incorporate this idea in the Markov chain of Figure 1.4 with the "updates to AI" state.

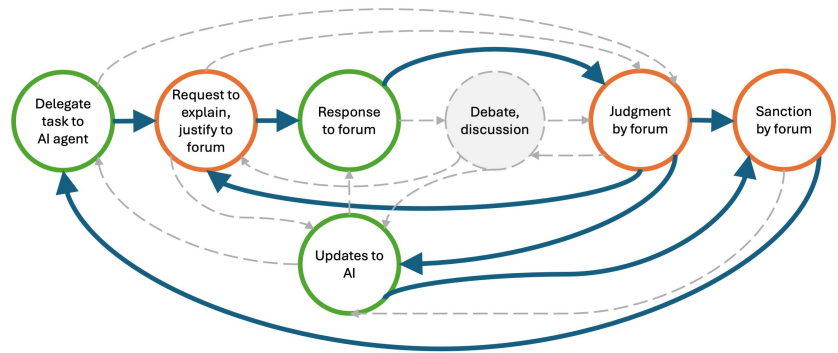

Figure 1.4 The states of the audit process for *pymetrics* represented using the Accountable AI Markov chain. The audit process implemented only some of the transitions between states, and also omitted the "Debate, discussion" state. This omitted state, and the transitions that were not implemented are grayed out in the figure.

### 1.4.2 Accountable AI gone wrong

As Sal Khan, the founder of Khan Academy, argues eloquently in his book "Brave new words,"[3] AI has the potential to revolutionize education by democratizing it: with AI every student could have a personal tutor at their fingertips (Khan, 2024). I certainly feel this to be true when I ask ChatGPT for help with high school biology material, so that I could discuss it with my kids. Khan even argues that, when it comes to evaluating the work of students, "AI can be auditable and accountable in ways that human[s] ... often aren't." His suggested path forward is as follows. First, we could train AI to avoid biases that humans might hold. Second, in many educational situations humans are not adequately held to account, so adding AI as "an extra check" could be helpful in making educators accountable to students.

---

[3] Ironically, the title of Khan's book is an homage to Aldous Huxley's "Brave New World," which is a dystopian story of a society divided into (of all things) intelligence-based castes (Huxley, 2022).

Yet, using AI, and more generally algorithms, to improve the accountability of educators, has been anything but problem-free. A good example is the use of a proprietary algorithm by the Houston Independent School District to evaluate teachers (Paige and Amrein-Beardsley, 2020; Wang, 2024; Cheong, 2024). Paige and Amrein-Beardsley (2020) provide a detailed account of how the school district used a proprietary, black-box, algorithm to "statistically link a teacher's contributions to students' growth on standardized tests, predominantly in mathematics and English/language arts, and hold teachers accountable for their students' collective growth over time or lack thereof." What is important from our perspective is this: neither the teachers, nor even the school district, could get information about how the proprietary software arrived at teacher evaluations. Yet, these evaluations were used to hold teachers accountable, including to decide on consequences, from terminations to merit-based pay raises. About a decade after the introduction of the algorithm, the Houston teachers won in court, and the use of the algorithm in this school district was discontinued. It is important for us to recognize how long it took for this reversal to happen and contrast it with the speed at which AI can be developed and deployed (Wang, 2024).

## 1.5  What's missing to make Accountable work in AI? What needs to change?

Let us return to the three constituent parts of accountability – a forum can request information from an agent about the actions of that agent, the forum and agent can discuss the information provided by the agent, and the forum can sanction the agent if the forum finds this appropriate. The sanction can be positive or negative. The forum can decide to sanction the agent either because of the way that the agent did (or did not) provide the requested information, or because of the content of the information. Note that for accountable AI, we need a socio-technical system for accountability. Next, let us look at what is missing to make accountability work in AI from two perspectives: first by considering the broad rules that will govern AI, and then by exploring some specific ideas for implementing accountable AI.

### 1.5.1 What's missing – the broad rules of engagement

To make the three constituent elements of accountability function for AI, we need clear governing structures that are enforced. Governing structures are being proposed and implemented at various scales. However, there is no broad agreement on which structures to implement, in part because there is no agreement on which values these governing structures should support. Furthermore, our ability to enforce any governing structure is wanting.

Let us start with governing structures that can ensure accountable AI. There is a sizable group of people who would agree with the argument put forth by Hoffman and Beato (2025) that AI needs "permissionless innovation." Hoffman and Beato argue against the need for developers of AI to get permission to deploy and operate their inventions. Instead they argue that a vaguely-defined accountability should ensure that AI functions to the benefit of humans. They cite Grossman (2015), who highlighted the permissionless participation of workers in ride sharing companies (in the role of drivers) and online marketplaces (in the role of sellers). He argued that the efforts of these workers are governed by "strict accountability" which in turn is made possible by the availability of information about their performance. The idea here is simple: while anyone can become a driver or seller, drivers with dirty cars, as well as sellers who do not fulfill orders, will get poor customer evaluations, and thus will not get customers in the future. Grossman was aware that "the stakes [can] get higher" if the activity in question is related to e.g. "transportation, finance and healthcare," but argued that we should still try to create "accountability-based models."

Hoffman and Beato (2025) show no such restraint. They argue for permissionless innovation for AI where the stakes would be very high indeed: they envision AI that can "help us reduce the threats of nuclear war, climate change, pandemics, resource depletion, and more." The accountability they propose appears to be that you should deploy a product, get feedback from customers, and then make appropriate changes to satisfy customers. This is truly a breathtaking argument. Here, customers play the role of the forum, and lack of adequate information about how the AI works, or poor AI performance, would presumably lead to negative sanctions by the customers in the form of refusal to work with the AI developer. This is an extremely forgiving system of accountability. It is like the King of England telling the barons: "We don't need the Magna Carta. Just trust me. And if you think I plunder too much, then move out of England!" This lack of true accountability simply does not match either the stakes (human lives and economic security), or the speed at which significant harm could be done with powerful AI technology[4]. It also relies on what Harari (2024) calls the "naive view of information" which is that more information always leads us to the truth (in this case to beneficial AI). But if we ever needed proof that this naive view is incorrect, we need look no further than today's social media.

The potential harm of AI, and the speed with which it might come, worries many. Harari (2024) warns that AI has the potential to be the equivalent of alien intelligence that will surpass humanity and might not be accountable to us. Mitchell et al. (2025) argue that "fully autonomous AI agents should not be built," which of course would eliminate the need for AI accountability. The EU AI Act bans AI applications that are classified as posing unacceptable risk (one example of banned AI is social ranking algorithms) (Thelisson and Verma, 2024). Suleyman (2023) expects that such prohibitions cannot be successful, and argues that we need to slow AI development to gain time for the development of governing structures for developing safe AI, and this includes developing structures for accountability. On the local level Raji et al. (2020) also argue for slowing down AI development by introducing internal algorithmic audits. We can think of the idea of slowing down AI development as incrementalism, which is an attitude of making small changes that add up over time, instead of attempting to quickly execute large, sweeping changes. As Berman and Fox (2023) argue, incrementalism can serve us well when we aim to make progress on a societal issue, while sudden, large shifts can have unintended consequences.

What needs to change then regarding rules of engagement is quite significant: humanity will need broader discussion and ultimately agreement on what we hope to accomplish with AI accountability. Do we want fast progress at the expense of safety, as in the case of permissionless innovation? Do we want to ban some technologies (the EU says yes)? Do we want incrementalism in AI development and deployment? And how do we develop our understanding and interpretation of accountability for AI for each of these scenarios, or the likely cacophony of different approaches?

### 1.5.2  What's missing – the socio-technical tools

The recent review of AI audits by Birhane et al. (2024) provides an insightful set of recommendations for those pursuing academic research on this topic. We already mentioned one: audits should focus on more than evaluation. However, one place where the methods of evaluation could lead to interesting progress is the exploration of the "debate, discussion" state in Figure 1.1.

---

[4]  What if the developer simply deploys the technology without implementing real accountability, as the billionaire in the novel "Termination Shock" by Stephenson (2021), who deploys (AI-based) geoengineering technology to reduce global temperatures?

Brandsma and Schillemans (2013), focusing on bureaucracies, point out the importance of (sometimes informal) discussions between the agent and the forum, saying that "[t]he principal may learn to articulate [their] preferences or start thinking about what is really at stake, and the agent may learn what its principal really wants to see accomplished." Yet, the authors found very few public administration-related studies that provide quantitative evaluations of discussions in the accountability process. However, such quantitative studies of the discussion phase of AI accountability could be very important, and particularly in the case of AI use by non-professional users (Janssen et al., 2019b). These users will increasingly engage with AI in a variety of situations (Mollick, 2024), including working while riding in a future automated vehicle (Teodor-ovicz et al., 2022), coding (Geyer et al., 2025), generating creative ideas (Shaer et al., 2024), and even supporting the work of courts (Solovey et al., 2025) . What will these users expect of AI in their ever-changing interactions? How will users differ in their expectations? How can interactions between users and AI be designed to support AI accountability? We expect that this is where supporting discussions can help – both discussions between designers and operators of AI and users (and more broadly stakeholders (San Vito et al., 2025)), as well as discussions between AI and its users. Enabling such discussions involves technological challenges, including representing the inner workings of the AI such that users can understand it (Gunning et al., 2019; Rudin, 2019), providing appropriate user interfaces (San Vito et al., 2025), carefully integrating AI that is procured from different developers (World Economic Forum, 2023), and providing information about the AI that is relevant to specific domains (e.g. healthcare (Miao et al., 2025) or automated driving (Janssen et al., 2019a).

Another approach that could support accountability is organizational – it would be to mimic the way that networking companies ensure interoperability. These companies create testing consortia, where an independent and neutral organization provides paid testing services. Such a service provider could create the expertise to assess accountability for a wide array of AI companies, and could potentially fulfill legal testing requirements as well.

Finally, as we think about technology for AI accountability, there are opportunities to learn from the way that accountability was established before the widespread use of AI. In this chapter we discussed many ideas about AI accountability that were originally developed with government and bureaucracy in mind. But there are other, perhaps less obvious, but very exciting connections waiting for us. One of those connections was discussed by Kate Darling, who argues that our relationship with robots (that is with embodied AIs) will often resemble our relationships with animals – thus, as we create accountable AI we can utilize humanity's extensive experiences with holding animals, and their owners, accountable for the actions of the animals (Darling, 2021). But more such connections are out there.